Localization and Perception for Control of a Low Speed Autonomous Shuttle in a

Campus Pilot Deployment

Thesis

Presented in Partial Fulfillment of the Requirements for the Degree Master of Science in

the Graduate School of The Ohio State University

By

Bowen Wen

Graduate Program in Electrical and Computer Engineering

The Ohio State University

2018

Thesis Committee

Levent Guvenc, Advisor

Rongjun Qin, Advisor

Abstract

Future SAE Level 4 and Level 5 autonomous vehicles will require novel applications of localization, perception, control and artificial intelligence technology in order to offer innovative and disruptive solutions to current mobility problems. Accurate localization is essential for self driving vehicle navigation in GPS inaccessible environments. This thesis concentrates on low speed autonomous shuttles that are mainly utilized for university campus intelligent transportation systems and presents initial results of ongoing work on developing solutions to the localization and perception challenges of a university planned pilot deployment orientated application. The paper treats autonomous driving with real time kinematics GPS (Global Positioning Systems) with an inertial measurement unit (IMU), combined with simultaneous localization and mapping (SLAM) with three-dimensional light detection and ranging (LIDAR) sensor, which provides solutions to scenarios where GPS is not available or a lower cost and hence lower accuracy GPS is desirable. The in-house automated low speed electric vehicle from the Automated Driving Lab is used in experimental evaluation and verification. An improved version of Hector SLAM was implemented on ROS and compared with high resolution GPS aided localization framework in the same hardware architecture. The overall configuration that combines ROS with DSpace controller can be easily transplantable prototype in other hardware architectures for future similar research. Real-world experiments that are reported here have been conducted in a small test area close to the Ohio State University AV pilot test route. are used for demonstrating the feasibility and robustness of this approach to developing and evaluating low speed autonomous shuttle localization and perception algorithms for control and decision making.

Acknowledgments

I would like to thank both my advisor, Levent Guvenc, Rongjun Qin, for providing me with the opportunity to work on an ambitious project, the full support for my research and the access to the most advanced facilities in the lab. Thank you also to Sukru Yaren and to other lab members who worked with me on this project and the technical help with the mechanical system setup. I would also like to thank the excellent Ohio State ECE faculties, from whom I have learned much. I would also like to acknowledge my parents and friends for their love and support.

Vita

2012..............................................................B.S. Power and Systems and Automation,

Xi'an Jiaotong University

2016..............................................................M.S. Electrical and Computer Engineering,

The Ohio State University

Publications

Wen, Bowen, Sukru Yaren Gelbal, Bilin Aksun Guvenc, and Levent Guvenc. Localization and Perception for Control and Decision Making of a Low Speed Autonomous Shuttle in a Campus Pilot Deployment. No. 2018-01-1182. SAE Technical Paper, 2018.

Fields of Study

Major Field: Electrical and Computer Engineering

Table of Contents

List of Tables

List of Figures

Chapter 1. Introduction

1.1 Intelligent Vehicle Development

In recent years, there has been increasingly enormous interests in making vehicles more intelligent and smart, both in the form of human-less autonomous vehicles and for advanced driver-assistance systems (ADAS) so as to improve the safety as well as the convenience for daily transportations. The Society of Automobile Engineers (SAE) defined five levels of autonomous driving, as summarized in Figure 1. Levels 1-3 require a licensed driver, but levels 4 and 5 allow driverless operation, which is necessary for many predicted benefits [1]. Accordingly, Autonomous vehicles are becoming a new piece of infrastructure attributed to the potential benefits. An increasingly larger academic research community is concentrating on this field and has contributed significantly to producing their prototype systems.

| SAE level | Name | Narrative Definition | Execution of Steering and Acceleration/ Deceleration | Monitoring of Driving Environment | Fallback Performance of Dynamic Driving Task | System Capability (Driving Modes) |
|---|---|---|---|---|---|---|
| *Human driver* monitors the driving environment | | | | | | |
| 0 | No Automation | the full-time performance by the *human driver* of all aspects of the *dynamic driving task*, even when enhanced by warning or intervention systems | Human driver | Human driver | Human driver | n/a |
| 1 | Driver Assistance | the *driving mode*-specific execution by a driver assistance system of either steering or acceleration/deceleration using information about the driving environment and with the expectation that the *human driver* perform all remaining aspects of the *dynamic driving task* | Human driver and system | Human driver | Human driver | Some driving modes |
| 2 | Partial Automation | the *driving mode*-specific execution by one or more driver assistance systems of both steering and acceleration/ deceleration using information about the driving environment and with the expectation that the *human driver* perform all remaining aspects of the *dynamic driving task* | System | Human driver | Human driver | Some driving modes |
| *Automated driving system* ("system") monitors the driving environment | | | | | | |
| 3 | Conditional Automation | the *driving mode*-specific performance by an *automated driving system* of all aspects of the dynamic driving task with the expectation that the *human driver* will respond appropriately to a *request to intervene* | System | System | Human driver | Some driving modes |
| 4 | High Automation | the *driving mode*-specific performance by an automated driving system of all aspects of the *dynamic driving task*, even if a *human driver* does not respond appropriately to a *request to intervene* | System | System | System | Some driving modes |
| 5 | Full Automation | the full-time performance by an *automated driving system* of all aspects of the *dynamic driving task* under all roadway and environmental conditions that can be managed by a *human driver* | System | System | System | All driving modes |

Figure 1 The SAE defined five vehicle automation levels [1]

Despite this trend, autonomous vehicles are not systematically organized and not friendly transplantable. Given that commercial vehicles protect their in-vehicle system interface from users, third-party vendors cannot easily test new components of autonomous vehicles [1]. In addition, sensors are not identical such that making it more difficult to share the same configuration among different platforms. More specifically, some low cost prototyped vehicles prefer to utilize limited resources of sensors, such as only low cost cameras, radars or GPS, whereas others might tend to use a combination of diverse sensors with more advanced and consequently more expensive configurations, including cameras, laser scanners, high resolution GPS receivers, and milli-wave radars. Additionally, regarding towards each of the sensor components, it tremendously depends on the specific capability and quality. For example, a 64 channel Velodyne LIDAR can demonstrate a dramatically different point cloud quality from that generated from a 16 channel Velodyne LIDAR and thus potentially influent the adaptivity and robustness of the platform. Although more comprehensively equipped autonomous vehicles, such as one typical configuration of advanced equipped autonomous car shown in Figure 2, naturally presents to have better performances and have already succeeded in autonomously driving through some urban challenging streets [2], such configurations are still not easily accessible to the public restricted by their high cost at the same time.

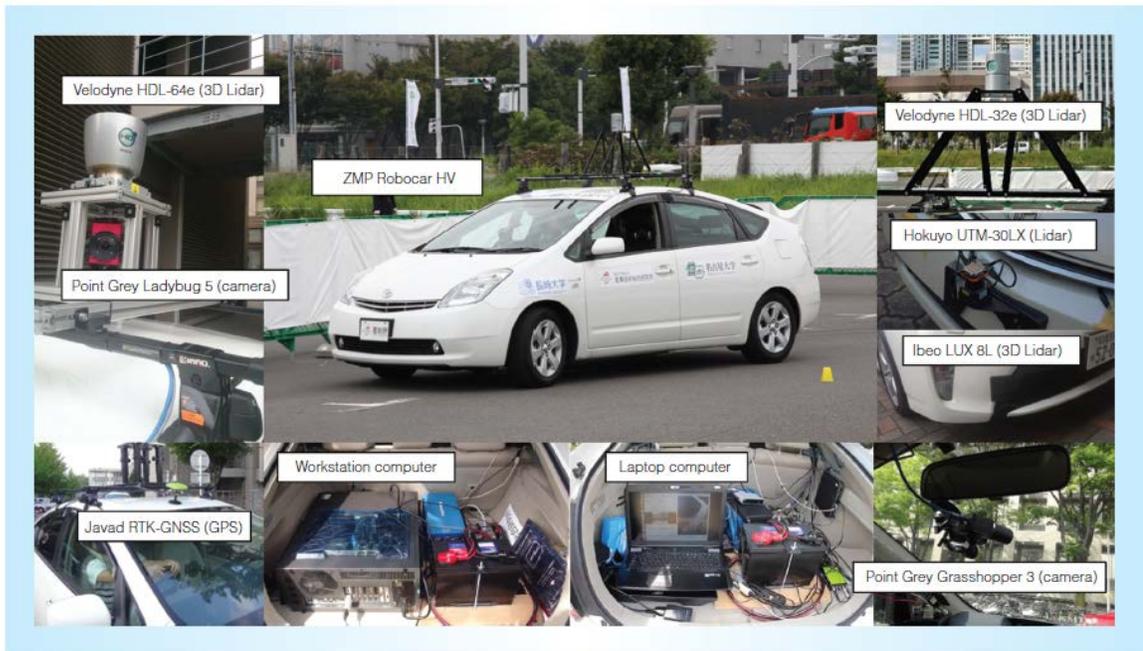

Figure 2 Example of hardware configuration of autonomous vehicle [1]

In addition to the issues related to hardware adaptivity, software problems of autonomous vehicles are simultaneously expected to be resolved. Taking into account the fact that an autonomous-vehicles platform is large scale, it is inefficient to build it up from scratch every time, especially for prototypes. Open-source software libraries are preferred for this purpose. Some libraries have been developed aiming to facilitate the multidisciplinary collaboration in research and development in the diverse technologies required by autonomous driving vehicles [3], where they introduced an open platform for autonomous vehicles that many researchers and developers can study to obtain a baseline for autonomous vehicles, design new algorithms, and test their performance, using a common interface. Their overall framework is depicted in Figure 3. SUMO [5] also introduced a open source traffic simulation package including net import and demand modeling components that described the state of the package as well as future developments and extensions, which helps to investigate several research topics e.g. route choice and traffic light algorithm or simulating vehicular communication.

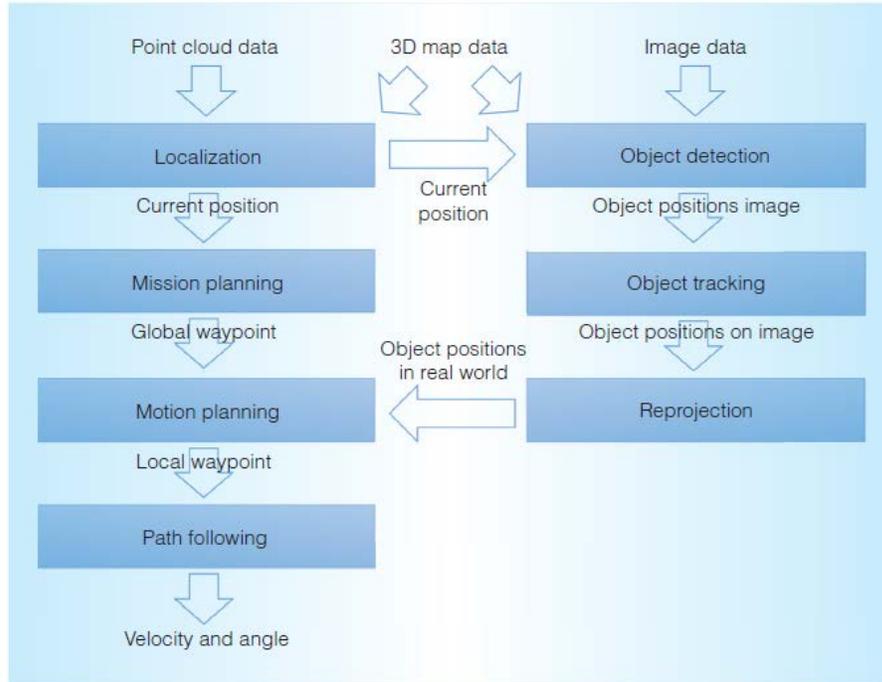

Figure 3 Basic control and dataflow of algorithms. The output velocity and angle are sent as commands to the vehicle controller [3]

However, while these libraries greatly benefinited the proceeding of autonomous driving research by providing general software framework for autonomous driving strategy simulation and experiment, specific software programs are highly correlated to the hardware components equipped on the specific autonomous vehicle. Low cost configurations of more economical sensors generally have higher demand for more complex and stable algorithms for compensation. Therefore, reliable and robust algorithms targeted towards low cost autonomous vehicles are highly desired for the commercialization of autonomous driving vehicles in the near future.

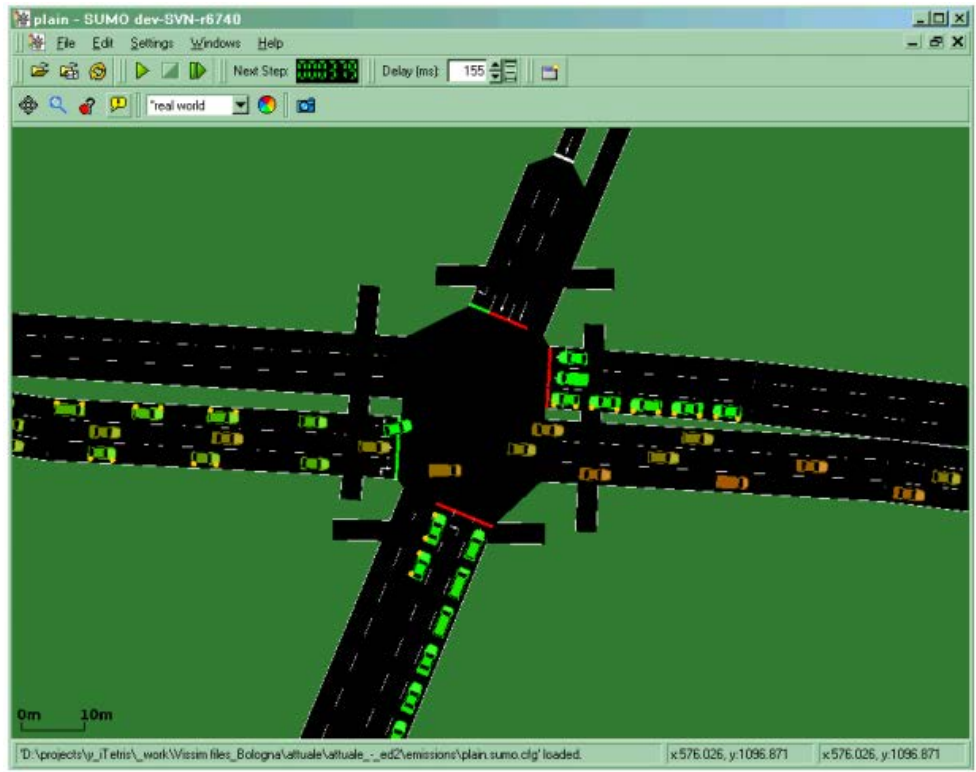

Figure 4 Screenshot of the graphical user interface coloring vehicles by

their CO2 emission in SUMO [5]

1.2 Vehicle Localization and Perception

As for the perception aspect of an intelligent autonomous vehicle, one of the most essential problem is the estimation of the vehicle position and orientation, which altogether are called pose, relative to some reference coordinate system or sometimes global coordinate system. Preferable possible reference coordinate systems for an autonomous driving vehicle can be the road beneath the vehicle that at the same time is also the starting position of the vehicle. While GPS is available or more generative global surrounding information is available, longitudinal or latitudinal coordinate representations or their converted scales are also a common approach for autonomous driving navigation system in urban areas.

Simultaneous localization and mapping (SLAM) as first proposed by Leonard and Durrant-Whyte [1] is used to build up maps of surrounding environment with the aid of sensors such as light detection and ranging (LIDAR) sensor or camera, while also estimating the position of a robot simultaneously as shown in Figure 5. A reliable and accurate solution of SLAM problems can implicitly lay the foundation for an autonomous navigation and control platform [6, 7].

The "solution" of the SLAM problem has been one of the notable successes of the robotics community over the past decade. SLAM has been formulated and solved as a theoretical problem in a number of different forms. SLAM has also been implemented in a number of different domains from indoor robots to outdoor, underwater, and airborne systems. At a theoretical and conceptual level, SLAM can now be considered a solved problem. However, substantial issues remain in practically realizing more general SLAM solutions and notably in building and using perceptually rich maps as part of a SLAM

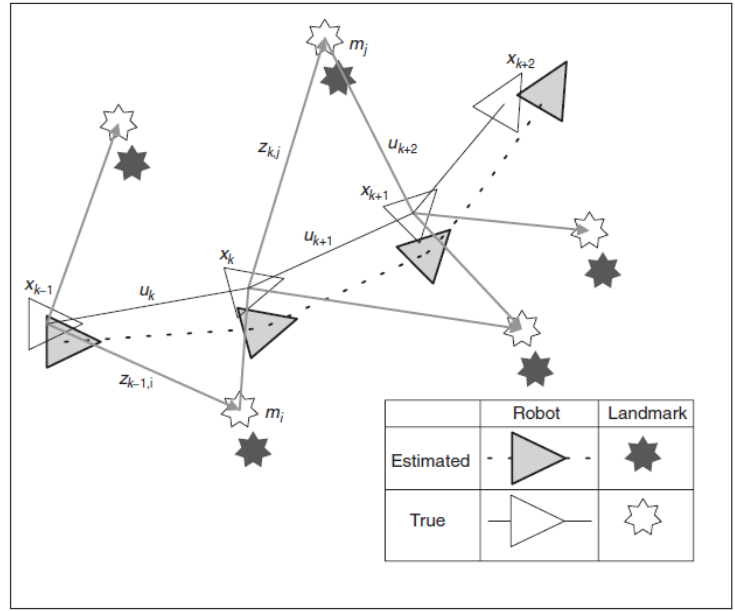

Figure 5 The essential SLAM problem. A simultaneous estimate of both robot and landmark locations is required. The true locations are never known or measured directly. Observations are made between true robot and landmark locations [20]

algorithm [20]. For its application to autonomous driving systems, the issues include mapping of static landmarks along with surrounding dynamic moving objects such as pedestrians, other vehicles, and bicyclists, which are especially common in urban environments. Large scale mapping also needs to be addressed both effectively and efficiently such that autonomous driving vehicles that move in large distance are still capable of accurately localizing itself. Although some existing SLAM algorithms to some extent ameliorate the problem of real time performance, the fact that vehicle localization in complex environments and driving at high speed is still challenging. Because consistent full solution to the combined localization and mapping problem would require a joint state composed of the vehicle pose and every landmark position, to be updated following each landmark observation. In turn, this would require the estimator to employ a huge state vector (on the order of the number of landmarks maintained in the map and exponentially grows as the environments becomes more complex) with computation scaling as about the square magnitude of the number of landmarks [20].

During the last decade, highly effective SLAM techniques have been developed and state-of-the-art two dimensional laser SLAM algorithms are now able to have satisfactory performance in terms of accuracy and computational speed (e.g. GMapping [8] and Hector SLAM [9]). In addition, researchers have successfully extended SLAM applicable scenarios from indoor environment to outdoor environment for autonomous vehicles [10, 11]. Probabilistic map distributions over environment properties followed by Bayesian inference in [12] increased robustness to environment variations and dynamic obstacles, which enabled the vehicle to autonomously drive for hundreds of miles in dense traffic on narrow urban roads. A fast implementation of incremental scan matching method based on occupancy grid map was introduced in [13] where data association was also applied to solve the multiple object tracking problem in a dynamic environment. Most of the previous work in the literature about SLAM methods has concentrated on the evaluation of localization performance whereas SLAM is utilized and evaluated as part of an automated path following system here in this study.

1.3 Thesis Objective and Scope

For the sake of development of smart city, the Ohio State University has designated a small segment in an underserved area of campus as an initial Autonomous Vehicle (AV) pilot test route for the deployment of SAE Level 4 low speed autonomous shuttles [24]. This thesis presents preliminary work towards proof-of-concept low speed autonomous shuttle deployment in this AV pilot test route which extends from the Automated Driving Lab through a 0.7 mile public road with a traffic light intersection and low speed traffic to our main research center. The approach is to develop and test elements of this autonomous system in the private parking lot right next to the Automated Driving Lab and in a realistic virtual replica of the AV pilot test route created within our Hardware-in-the-Loop (HiL) simulator environment. This study concentrates on LIDAR SLAM based localization for path tracking, a simple decision making logic for automated driving and experimental and simulation results.

Chapter 2.   2D LIDAR SLAM algorithm

The SLAM based localization algorithm is presented in this section. In this study, ground plane is always assumed to be flat and hence only 2D mapping and localization are required while z direction pose information in Cartesian coordinate system is not necessarily considered. In the following algorithm, the pose state vector $(x, y, \theta)^T$, comprised of 2D Cartesian coordinates and orientation angle, and thus three degrees of freedom (DOF), is used to represent the pose information for the low speed autonomous shuttle. As has been presented, the 16 channel Velodyne LIDAR can provide 3D point cloud including 360 degree FoV information of the surrounding environment. However, in this context, considering the constraint of the processor in this configuration, additional computational complexity will negatively affect the whole system in terms of real time performance. Therefore, so as to obtain planar scan information, 3D point cloud is projected into 2D space.

Before the projection, ground plane as seen in Figure 6 needs to be removed by building up occupancy height map (Section 2.2.1). Once the planar scan end points are obtained, scan matching process is used to align the current scan end points either to those in last frame or to the built up map in order to derive the pose transformation of the shuttle. A more reliable and accurate optimization framework inspired by Hector SLAM [9]  is imposed for the scan matching process, where more reasonable stop criteria is also introduced (section 2.2.2).

2.1 Ground plane Removal and Projection

Occupancy height map is built up for ground plane removal. The LIDAR position is selected as the origin and the Cartesian coordinate system is built with the *x-y* plane representing the ground plane and the z axis being vertical to it. As shown in Figure 7, from a top-down view, we divide the *x-y* plane into many square cells of equal size. In this work, cell size is set to 0.2m x 0.2m. For each of the 3D points $P_i=(x_i,y_i,z_i)^T$, we can find a cell *Cj* that it belongs to. Subsequently, for each of the cells *Cj* by comparing the heights of the points to a threshold $h_{thres}$ (set to 0.3m in this work), if

$$z_{max,j} - z_{min,j} \leq h_{thres}$$

(1)

then this cell is defined as not occupied or comprised of ground plane and thus left as empty. If

$$z_{max,j} - z_{min,j} \geq h_{thres}$$

(2)

then this cell is defined as occupied and all the 3D points included in it are remained for further projection.

In the projection step, polar coordinate system is used to represent the position of each scan end point in 2D plane. For each 3D point $P_i$, its angular position in *x-y* plane can be expressed as:

$$\alpha_i = atan2(y_i, x_i)$$

(3)

where $atan2$ is four-quadrant inverse tangent and hence $\alpha_i \in [-\pi, \pi]$. The range of the 2D scan corresponding to the 3D point $P_i$ can be expressed as:

$$range_i = \sqrt{x_i^2 + y_i^2}$$

(4)

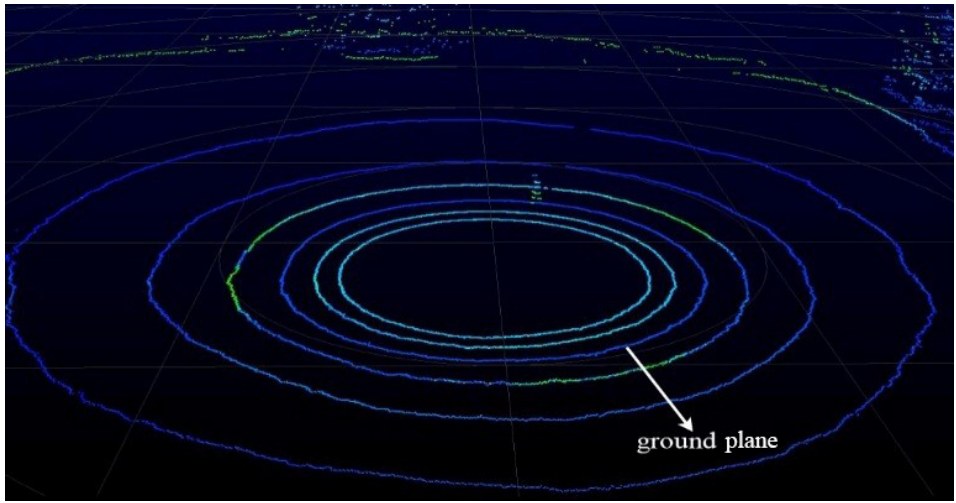

Figure 6 Raw 3D point cloud with ground plane

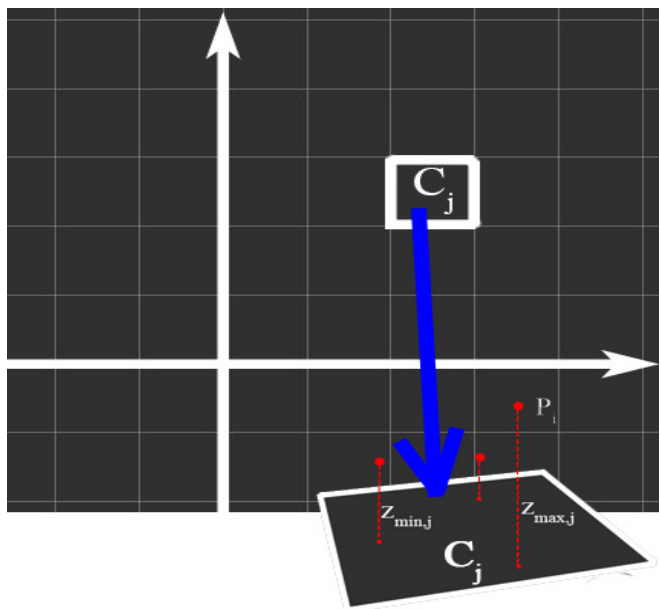

Figure 7 Occupancy height map. Cj is one of the cells. Height of every cell is determined by the maximum height difference in that cell

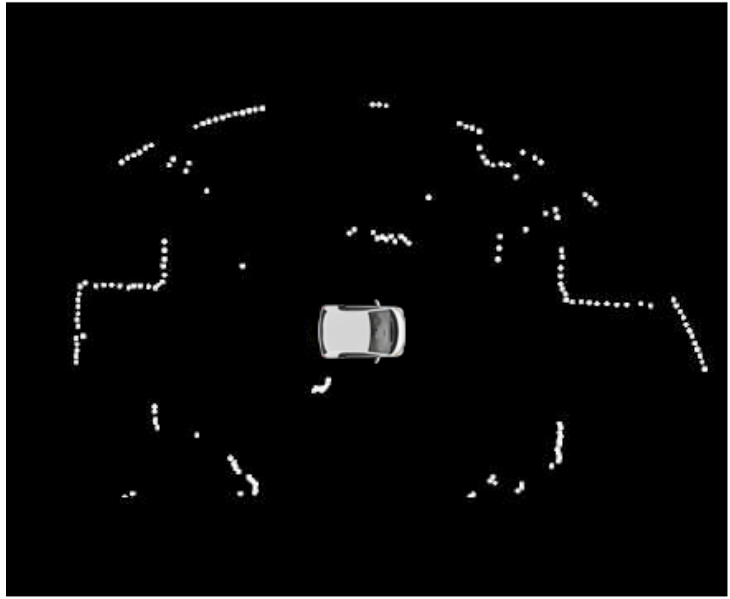

Figure 8 Projected 2D scan end points from surrounding 3D point cloud

Note that there can be more than one projected 2D scan point in the same direction with different ranges. The ultimate range of 2D scan end point is the smallest range in that direction. Therefore, every projected 2D scan beams with their associated scan end points can be identified by angular positions, as shown in Figure 8. The corresponding pseudo-code of ground plane removal and 3D point cloud projection are provided in Algorithm 1 and Algorithm 2 respectively.

| Algorithm: ground_noise_removal($P$, $h_{thres}$) |
|---|
| Initialize grids $C$ in *x-y* plane |
| Initialize non-ground point cloud $P' \leftarrow \emptyset$ |
| For each ($P_i \in P$) do |
|     $j \leftarrow$ grid index $P_i$ belongs to |
|     $C_j \leftarrow C_j \cup \{P_i\}$ |
|     $z_{Cj} \leftarrow z_{Cj} \cup \{z_{Pi}\}$ |
| For each ($C_j \in C$) do |
|     $z_{Cj,max} \leftarrow argmax\{z_{Cj}\}$ |
|     $z_{Cj,min} \leftarrow argmin\{z_{Cj}\}$ |
|     If $\lvert z_{Cj,max} - z_{Cj,min} \rvert \geq h_{thres}$ then |
|         $P' \leftarrow P' \cup \{\text{points in } C_j\}$ |
| Return $P'$ |

Algorithm 1 Ground plane removal for 3D point cloud

```
Algorithm: point_cloud_projection(P)

Initialize ranges r as large values in every direction

Initialize scan end points P'

For each (P_i ∈ P) do

    α_i ← atan2(x_i, y_i)

    r_i ← √(x_i² + y_i²)

    If r_i < r_{α_i} then

        r_{α_i} ← r_i

        P' ← P' ∪ {r_{α_i}, α_i}

Return P'
```

Algorithm 2 3D point cloud projection to 2D scan end points

## 2.2 Map Generation

To be able to represent arbitrary environments, an occupancy grid map (shown in Figure 9) is used, which is a proven approach for mobile robot localization using LIDARs in real-world environments [18]. However, because of the continuous states in real world, discrete property from the traditional occupancy grid map is unable to effectively represent the locations that are inside the grids, which naturally restricted the precision of real corresponding occupancy value of that location and thus negatively influenced the accuracy of following procedures such as scan alignment as well as pose estimations. In addition, discrete nature of such occupancy grid map does not allow its computations related to derivations, which can be a significant problem for optimization based scan matching alignment.

Therefore, in this study, the same map generation and representation approach was adopted from Hector SLAM [9], where an interpolation scheme allowing sub-grid cell accuracy through bilinear interpolation is employed for both estimating occupancy probabilities and derivatives. Intuitively, the grid map cell values can be viewed as samples of an underlying continuous probability distribution as shown in Figure 10, where $P_{00}$, $P_{10}$, $P_{01}$, $P_{11}$ are vertices of one occupancy grid sample and $\frac{\partial M}{\partial x}, \frac{\partial M}{\partial y}$ are the gradients of occupancy value with respect to the coordinates. The occupancy values $M$ in the grids are in the range of [0,1] and any grids whose occupancy probability higher than a threshold $\pi$ (set as 0.5 in this work) is regarded as occupied in the map. $x, y$ is the coordinate of one interested scan end point location and point $P_m$ is the located point whose occupancy value has been interpolated through bilinear interpolation.

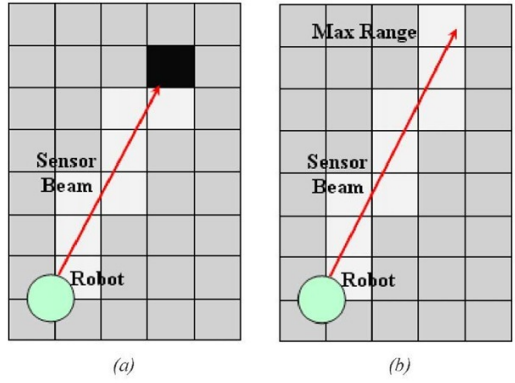

Figure 9 Occupancy grid map representation for mobile robot navigation. Color represents the confidence of occupancy [18]

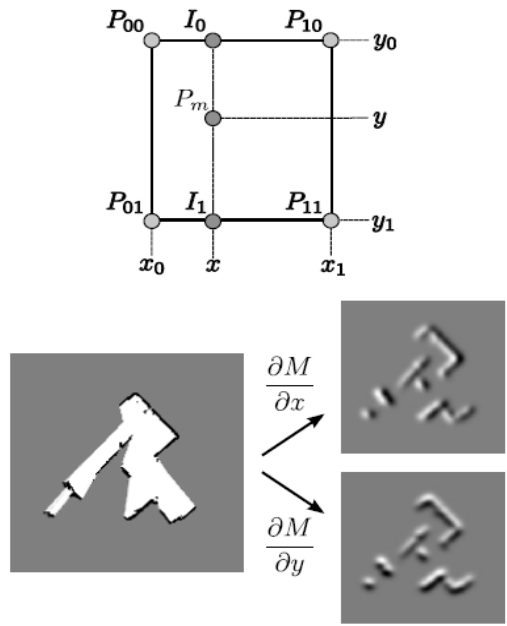

Figure 10 Bilinear interpolation of the occupancy grid map. Point Pm is the

point whose value has been interpolated [9]

Once we have the continuous coordinate representation of an interested scan end point $P_m$ through rigid transformation in this generated map coordinate system, we are now able to approximate the occupancy value $M(P_m)$ as well as the gradient $\nabla M(P_m) = (\frac{\partial M(P_m)}{\partial x}, \frac{\partial M(P_m)}{\partial y})$ by using the four neighboring vertices of the grid $P_{00}, P_{10}, P_{01}, P_{11}$ by the linear interpolation as follows:

$$M(P_m) \approx \frac{y - y_0}{y_1 - y_0}(\frac{x - x_0}{x_1 - x_0}M(P_{11}) + \frac{x_1 - x}{x_1 - x_0}M(P_{01})) +$$
$$\frac{y_1 - y}{y_1 - y_0}(\frac{x - x_0}{x_1 - x_0}M(P_{10}) + \frac{x_1 - x}{x_1 - x_0}M(P_{00}))$$

(5)

$$\frac{\partial M}{\partial x}(P_m) \approx \frac{y - y_0}{y_1 - y_0}(M(P_{11}) - M(P_{10})) + \frac{y_1 - y}{y_1 - y_0}(M(P_{01}) - M(P_{00}))$$

$$\frac{\partial M}{\partial y}(P_m) \approx \frac{x - x_0}{x_1 - x_0}(M(P_{11}) - M(P_{10})) + \frac{x_1 - x}{x_1 - x_0}(M(P_{01}) - M(P_{00}))$$

(6)

Surrounding environments of driving scenarios in the real world contain objects of many sizes, and these objects contain features of many sizes. As a result, the scan matching procedure that is applied only at a single scale occupancy grid map may miss information at other scales. Moreover, single scale high resolution occupancy grid map can ameliorate the problem of precision but inevitably introduce higher computational complexity, requiring more computation time, which is not preferred for a real time system for autonomous driving vehicles. Such problems can be mitigated by using a multi-resolution map representation similar to image pyramid approaches used in computer vision and image processing [22]. Following the same strategy in Hector SLAM [9], a multiple occupancy grid maps with each coarser map having half the resolution of the preceding one was employed in this work. However, the multiple map levels are not generated from a single high resolution map by applying Gaussian filtering and down-sampling as is commonly done in image processing. Instead, different maps are kept in memory and simultaneously updated using the pose estimates generated by the alignment process. This generative approach ensures that maps are consistent across scales while at the same time avoiding costly down-sampling operations. The scan alignment process is started at the coarsest map level, with the resulting estimated pose getting used as the start estimate for the next level, similar to the approach presented in [23]. This multi-level pose estimation from coarse to refinement also dramatically boosted the pose estimation speed by shrinking the optimization exploration space for every next occupancy grid map level. Another positive side-effect is that the immediate availability of different levels of resolutions of the occupancy grid maps which can be optionally selected to adapt to different navigation task requirements.

Figure 11 shows an example of the generated occupancy grid map from our collected data. (a) shows the single scale resolution representation of occupancy grid map and (b) shows the overlapped 5 level multi-resolution pyramid representation of occupancy grid map. As can be observed from the two generated maps, the multi-level resolution map demonstrated more concrete discriminatory outline of the occupied grids, which implicitly provides higher precision occupancy values for estimating poses in the following scan matching step.

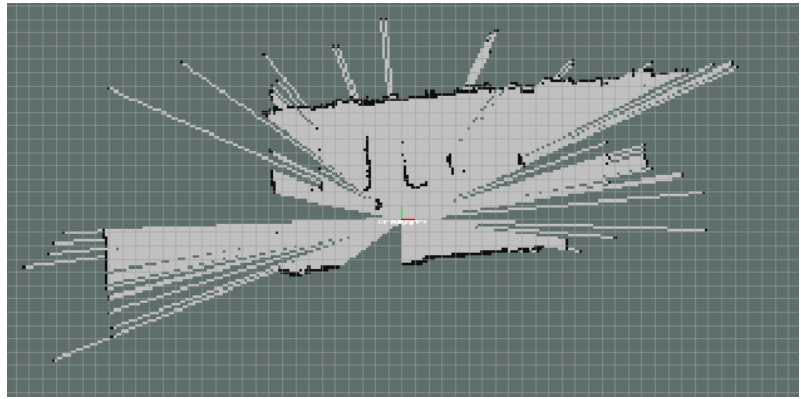

(a)

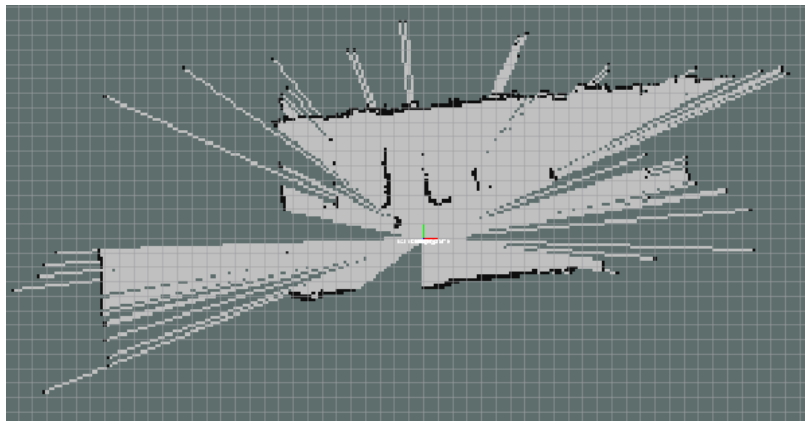

(b)

Figure 11 (a) Single scale resolution representation of occupancy grid map. (b) overlapped 5 level multi-resolution pyramid representation of occupancy grid map.

## 2.3 Scan Matching

Due to the high accuracy and frequency of modern LIDAR, iterative optimization algorithms are now possible to minimize the error between obtained scan end points and built up maps, delivering the optimal alignment in the scan matching step. In this work, instead of Gauss-Newton optimization performed in Hector SLAM [9], the Levenberg–Marquardt algorithm [16] is applied to provide faster convergence for same accuracy compared with Gauss-Newton optimization, which can tremendously benefit the real time system on autonomous shuttles. Given the generated map occupancy value $M(P_m)$ corresponding to the continuous map point location $P_m = (x_m, y_m)^T$, our goal is to find the rigid transformation $\xi = (p_x, p_y, \theta)^T$ which minimizes the overall summation of occupancy error between the current scan end points and the most updated map. Given that the generated occupancy value of the current scan end point location is $M(P_m)$ while the supposed occupancy value of the current scan end point location should be 1 based on the observation of the high frequency accurate laser scanner, consequently the objective function and desired rigid transformation can be defined as:

$$E = \min \sum_{i=1}^{n} [1 - M(S_i(\xi))]^2$$

(7)

$$\xi^* = \arg\min_{\xi^*} \sum_{i=1}^{n} [1 - M(S_i(\xi))]^2$$

(8)

where $n$ is the number of scan end points, $s_i = (s_{i,x}, s_{i,y})^T$ is the world coordinate of the transformed scan end point. $S_i(\xi)$ is a function of $\xi$ that transforms scan end point coordinate into world system, expressed as:

$$S_i(\xi) = \begin{pmatrix} \cos(\theta) & -\sin(\theta) \\ \sin(\theta) & \cos(\theta) \end{pmatrix} \begin{pmatrix} s_{i,x} \\ s_{i,y} \end{pmatrix} + \begin{pmatrix} p_x \\ p_y \end{pmatrix}$$

(9)

and $M(S_i(\xi)) \in [0,1]$ is the occupancy value at the location given by $S_i(\xi)$. Once this is performed, the optimal transformation that best aligns the current frame with the most updated map points is obtained.

This quadratic cost function $E$ can be solved by Levenberg–Marquardt algorithm [16] efficiently. Starting from an initial estimation of the transformation, e.g. the optimal transformation provided in last frame, $\xi_0$, in every iteration, a transformation update $\Delta\xi$ is added to the accumulated transformation so far, $\xi$, so as to move forward to the minimum point and further minimize the function. Intuitively, by each iteration step, the cost function is closer to 0:

$$E = \sum_{i=1}^{n}[1 - M(S_i(\xi + \Delta\xi))]^2 \rightarrow 0$$

(10)

By replacing $M(S_i(\xi + \Delta\xi))$ with its Taylor series expansion, we obtain:

$$E \approx \sum_{i=1}^{n}[1-M(S_i(\xi))-\nabla M(S_i(\xi))\frac{\partial S_i(\xi)}{\partial \xi}\Delta\xi]^2 \to 0$$

(11)

By letting the partial derivative with respect to *Δξ* equal to 0:

$$\frac{\partial E}{\partial(\Delta\xi)} = 2\sum_{i=1}^{n}[\nabla M(S_i(\xi))\frac{\partial S_i(\xi)}{\partial \xi}]^T \cdot ...$$

$$\cdot \sum_{i=1}^{n}[1-M(S_i(\xi))-\nabla M(S_i(\xi))\frac{\partial S_i(\xi)}{\partial \xi}\Delta\xi]=0$$

(12)

According to Levenberg–Marquardt algorithm, the optimal solution for *Δξ* can be determined by:

$$\Delta\xi = (H^{-1}+\lambda I)\sum_{i=1}^{n}w_i \cdot [\nabla M(S_i(\xi))\frac{\partial S_i(\xi)}{\partial \xi}]^T[1-M(S_i(\xi))]$$

(13)

where $w_i$ is weight associated with point $P_i$, which mainly down weights the low quality scan end points with big error and hence enhance robustness against noise [18]. $\lambda$ is a damping parameter (initially set to 0.01 in this work), *I* is identity matrix, *H* is weighted approximate Hessian matrix, defined by:

$$H = \sum_i w_i \cdot [\nabla M(S_i(\xi)) \frac{\partial S_i(\xi)}{\partial \xi}]^T [\nabla M(S_i(\xi)) \frac{\partial S_i(\xi)}{\partial \xi}]$$

(14)

From equation (9), we can easily calculate the derivative of the transformed scan end point with respect to the rigid transformation $\frac{\partial S_i(\xi)}{\partial \xi}$ as:

$$\frac{\partial S_i(\xi)}{\partial \xi} = \begin{bmatrix} \frac{\partial S_{i,x}(\xi)}{\partial p_x} & \frac{\partial S_{i,x}(\xi)}{\partial p_y} & \frac{\partial S_{i,x}(\xi)}{\partial \theta} \\ \frac{\partial S_{i,y}(\xi)}{\partial p_x} & \frac{\partial S_{i,y}(\xi)}{\partial p_y} & \frac{\partial S_{i,y}(\xi)}{\partial \theta} \end{bmatrix}$$

(15)

$$\frac{\partial S_{i,x}(\xi)}{\partial p_x} = 1, \quad \frac{\partial S_{i,x}(\xi)}{\partial p_y} = 1, \quad \frac{\partial S_{i,x}(\xi)}{\partial \theta} = -s_{i,x} \sin\theta - s_{i,y} \cos\theta$$

$$\frac{\partial S_{i,y}(\xi)}{\partial p_x} = 1, \quad \frac{\partial S_{i,y}(\xi)}{\partial p_y} = 1, \quad \frac{\partial S_{i,y}(\xi)}{\partial \theta} = s_{i,x} \sin\theta + s_{i,y} \cos\theta$$

(16)

By replacing equation (15) with equations (16), we can obtain the direct relationship of the derivatives:

$$\frac{\partial S_i(\xi)}{\partial \xi} = \begin{bmatrix} 1 & 1 & -s_{i,x} \sin\theta - s_{i,y} \cos\theta \\ 1 & 1 & s_{i,x} \sin\theta + s_{i,y} \cos\theta \end{bmatrix}$$

From the gradient of map occupancy value with respect to a scan end point location *(x,y)* as discussed in equation (6), a more concrete representation of $\nabla M(P_m)$ can be written as:

$$\nabla M(P_m) = \left[ \frac{\partial M(P_m)}{\partial x_m} \quad \frac{\partial M(P_m)}{\partial y_m} \right]$$

(17)

$$\nabla M(P_m) = \left[ \begin{array}{cc} \frac{y-y_0}{y_1-y_0}(M(P_{11})-M(P_{10})) + & \frac{x-x_0}{x_1-x_0}(M(P_{11})-M(P_{10})) + \\ \frac{y_1-y}{y_1-y_0}(M(P_{01})-M(P_{00})) & \frac{x_1-x}{x_1-x_0}(M(P_{01})-M(P_{00})) \end{array} \right]$$

(18)

By solving $\Delta \xi$, $\xi$ is updated by:

$$\xi \leftarrow \xi + \Delta \xi$$

(19)

and that makes $\xi$ iteratively move forward to the optimal transformation $\xi^*$.

Consider the scenario illustrated in Figure 12, which occurs frequently to a laser scanner sensor in the real world autonomous driving experiment. The top image shows the state of the mobile robot from last frame and the bottom image shows the mobile robot in the current frame. The black line in both images depicts an example of the same oriented laser scan from the mobile robot. In the bottom image, the black dotted line depicts the ideal transformed laser scan from the current frame to last frame. It can be intuitively observed that by naively following the above mentioned framework, even if we assume that we are able to obtain the real rigid transformation $\xi^*$ from the current frame to the last frame and apply it to the current frame, the transformed scan end point is not guaranteed to be located in the grid with the correct occupancy value because of the occlusion by the obstacle.

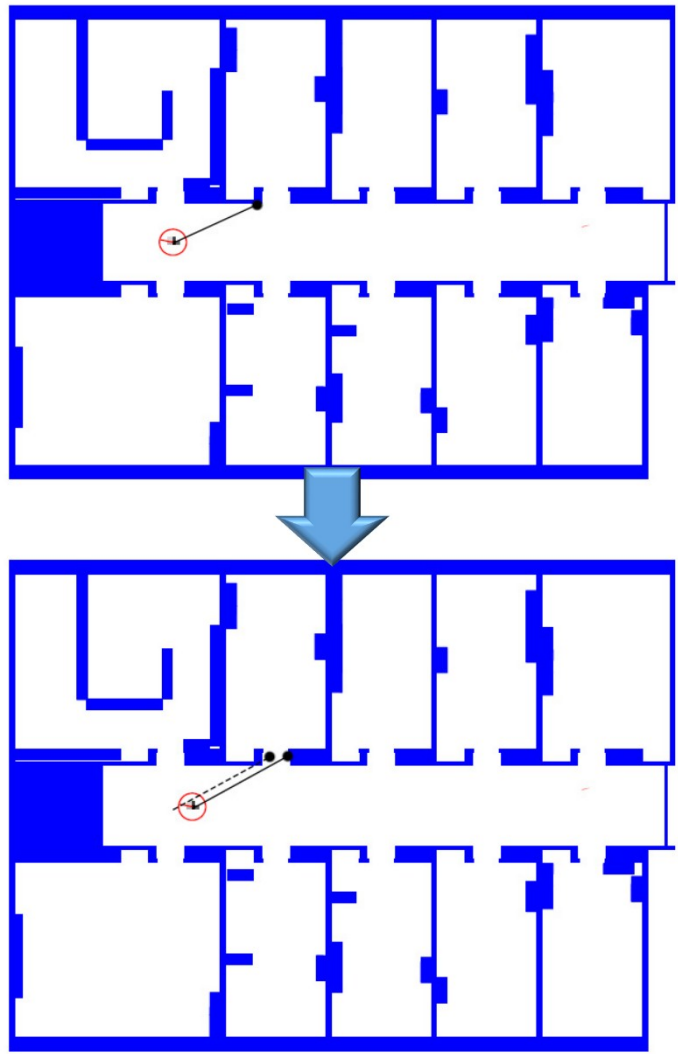

Figure 12 Top: mobile robot position of last frame. Bottom: mobile robot in current frame, where the black dotted line depicts the ideal transformed laser scan from the current frame to last frame

As a result, in this study which is based on iterative optimization to minimize the loss function, such scan end points should be regarded as outliers and not contribute to the total loss. In equation (14), unlike the original Hector SLAM where total losses are contributed by all laser scan end points, in this work, we also proposed a weight coefficient $w_i$, utilized to down-sample the contribution of the outlier scan end points, to represent the modified robust Huber penalties depending on its multiplied loss value *a* by ignoring the outlier contributions and the maximum loss is expected be the same as occupancy probability threshold $\pi$ for outliers. It can be then written as:

$$w_i = \begin{cases} a^2 & if \ |a| \leq \pi \\ \pi^2 & otherwise \end{cases}$$

(20)

Figure 13 shows the illustrated comparison between the original loss value and the weighted loss value with respect to the residual.

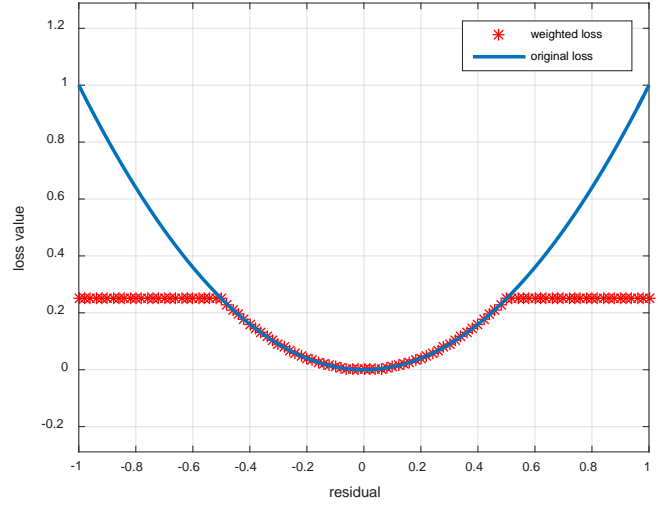

Figure 13 comparison between the original loss value and the weighted loss value with respect to the residual value ($\pi = 0.5$).

Moreover, In contrast to the practical implementation in Hector SLAM [9], where fixed iteration step setting is employed to evaluate the Gauss-Newton optimization, in addition to setting a maximum iteration step (10 in this work), we hereby proposed a more reasonable stop condition before reaching the maximum iteration step, which has been proven to ensure sufficient convergence while avoiding unnecessary iterations caused by oscillation around the optimal solution:

$$\|\Delta \xi\| < \varepsilon$$

(21)

where operator $\|\cdot\|$ denotes Frobenius norm, $\varepsilon$ is a parameter for threshold and is set to 0.001 in this work. $E_k$ is the cost function in the $k^{th}$ iteration step.

Chapter 3. Model Setup

3.1 Vehicle Dynamics

The vehicle model and path following algorithm used are presented briefly in this and the following section. The lateral dynamics and path tracking error model is illustrated in Figure 14 and given in state space form as:

$$\begin{bmatrix} \dot{\beta} \\ \dot{r} \\ \Delta\dot{\psi} \\ \dot{y} \end{bmatrix} = \begin{bmatrix} a_{11} & a_{12} & 0 & 0 \\ a_{21} & a_{22} & 0 & 0 \\ 0 & 1 & 0 & 0 \\ V & l_s & V & 0 \end{bmatrix} \begin{bmatrix} \beta \\ r \\ \Delta\psi \\ y \end{bmatrix} + \begin{bmatrix} b_{11} & 0 \\ b_{21} & 0 \\ 0 & -V \\ 0 & l_s V \end{bmatrix} \begin{bmatrix} \delta_f \\ \rho_{ref} \end{bmatrix}$$

(22)

where $\beta$ is side slip angle, $r$ is yaw rate, $V$ is combination of lateral and longitudinal velocity of the vehicle body, $\Delta\Psi$ is yaw angle relative to the tangent of the desired path, $l_s$ is the preview distance and $y$ is lateral deviation from desired path with respect to preview distance. The control input is the steering angle $\delta_f$. $\rho_{ref}=1/R$ is the road curvature where $R$ is the road radius. Other terms in the state space model are:

$$a_{11} = -\frac{C_r + C_f}{mV}, a_{12} = -1 + \frac{C_r l_r - C_f l_f}{mV^2}$$

$$a_{21} = -\frac{C_r l_r - C_f l_f}{J}, a_{22} = -\frac{C_r l_r^2 - C_f l_f^2}{JV^2}$$

$$b_{11} = \frac{C_f}{mV}, b_{12} = \frac{C_f l_f}{J}$$

(23)

where $m$ is the vehicle mass, $J$ is the moment of inertia, $\mu$ is the road friction coefficient, $C_f$ and $C_r$ are the cornering stiffnesses, $l_f$ is the distance from the Center of Gravity of the vehicle (CG) to the front axle and $l_r$ is the distance from the CG to the rear axle.

3.2 Path Tracking

The low level automated driving tasks are lateral and longitudinal control. The path determination and path tracking error computation are described briefly in this section. The path tracking model consists of two parts, which are offline generation of the path and online calculation of the error according to the generated path. These parts are explained in following subsections.

3.2.A. Offline Path Generation

The path following algorithm employs a pre-determined path to be provided to the autonomous vehicle to follow [15]. This map is generated from GPS waypoints where these points can be pulled from an online map or can be collected through recording during a priori manual driving. These data points are then divided into smaller groups named segments with equal number of data points for ease of formulation. These segments are both used for curve fitting and velocity profiling through the route. After dividing the road into segments, a process of fitting a third order polynomial is performed as:

$$X_i(\lambda) = a_{xi}\lambda^3 + b_{xi}\lambda^2 + c_{xi}\lambda + d_{xi}$$
$$Y_i(\lambda) = a_{yi}\lambda^3 + b_{yi}\lambda^2 + c_{yi}\lambda + d_{yi}$$

(24)

Where *i* represents the segment number and terms *a, b, c, d* are polynomial fit coefficients for the corresponding segment. Fitting the data points provides effective replication of the curvature that the road carries and also eliminates the noise in the GPS data points. To provide a smooth transition from one segment to another by satisfying continuity of the polynomials and their first derivatives in *X* and *Y,* we use:

$$X_i(1) = X_{i+1}(0)$$
$$Y_i(1) = Y_{i+1}(0)$$

(25)

The *X* and the *Y* points derived from the GPS latitude and longitude data using a degree to meter conversion, are fit using a single parameter $\lambda$, where $\lambda$ is the variable for the fit which varies across each segment between 0 and 1, resulting in:

$$\frac{dX_i(1)}{d\lambda} = \frac{dX_{i+1}(0)}{d\lambda}$$

$$\frac{dY_i(1)}{d\lambda} = \frac{dY_{i+1}(0)}{d\lambda}$$

(26)

### 3.2.B. Error Calculation

After the generation of path coefficients, an error is calculated for the lateral controller to use as input. Heading and position of the vehicle is provided by means of localization, in this case either SLAM or GPS. Using these, the location of the car with respect to the path in other words the deviation from the path is calculated. This approach reduces both oscillations and steady state lateral deviation compared to calculation with respect to position only. In order to find an equivalent distance parameter to add to the first component distance error, a preview distance $l_s$ is defined. Then, the error becomes:

$$y = h + l_s \sin(\Delta\psi)$$

(27)

Where $\Delta\psi$ is the net angular difference of heading of the vehicle from the heading tangent to the desired path and $y$ is the total error of the vehicle computed at preview distance $l_s$ as is illustrated in Figure 15.

Finally, error is fed to a robust PID controller which controls the actuation of steering of the vehicle.

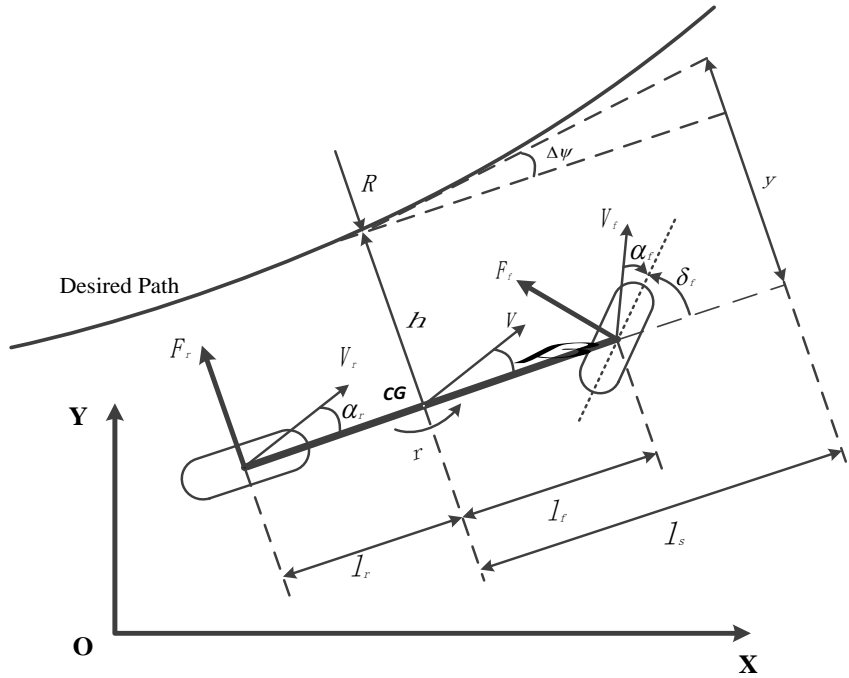

Figure 14 Illustration of single track model

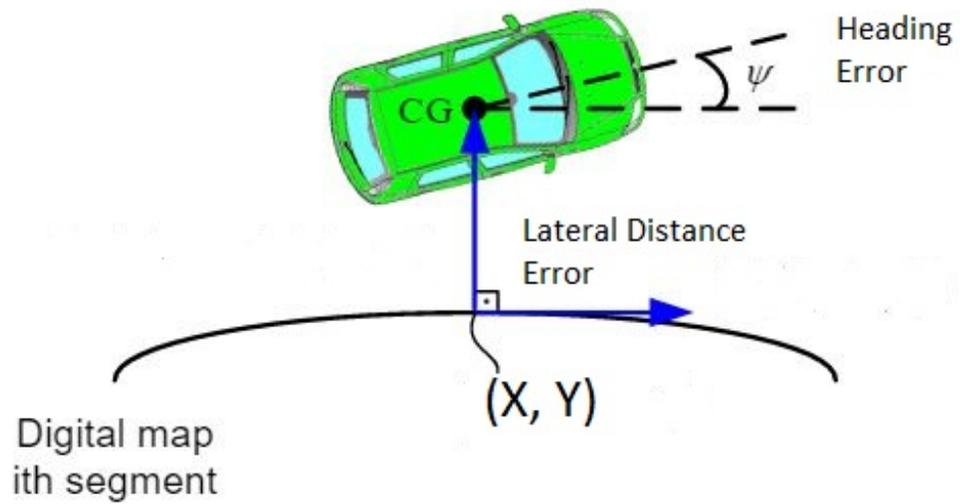

Figure 15 Illustration of error calculation

3.3 Hardware and Platform

The vehicle used in the experiments for this study is a small, low speed, fully-electric two seater shuttle used for ride sharing applications (Dash EV) as shown in Figure 16. The architecture and hardware presented in this paper is general in nature and also implemented on other vehicles in our lab [14]. In order to achieve autonomous driving capability, steering, throttle and brake in this vehicle were converted to by-wire. This is done by adding actuators into the vehicle, since it was not built with them as some of the commercial sedan vehicles. For steering actuation, a smart motor was connected to the steering mechanism through gears. For brake actuation, a linear electric motor was fixed behind the brake pedal, that pushes or pulls according to the position command. For throttle, an electronic by-pass circuit was constructed and used to override the throttle signal that is sent to vehicle Electronic Control Unit (ECU) with the throttle command.

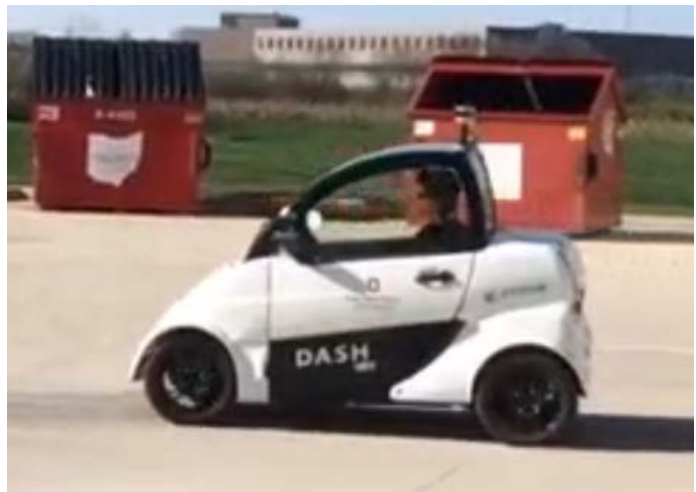

Figure 16 Dash EV, our experimental vehicle

Sensors are added for localization and environmental perception after steering, throttle and brake functions are converted to drive-by-wire. These sensors are GPS, a LIDAR sensor, a Leddar sensor (shown in Figure 17) and a Point Grey camera used in this paper as a backup sensor. The Leddar sensor is a solid-state LIDAR which we use to get information about the obstacles in front of the vehicle. These obstacles can be vehicles, pedestrians, bicyclists etc. It is mainly used for emergency purposes, when there is an obstacle very close to the vehicle which creates a need to stop. It can be also used in low speed car following applications such as Adaptive Cruise Control (ACC) since its range is 50 m. For localization, GPS and LIDAR sensors were used. We use OXTS XNAV550, a differential GPS with Real-Time Kinematic (RTK) correction capability shown in Figure 19, which provides about 2-5 cm accuracy when RTK correction signals are used. Also with the differential antennas, it provides heading information even while the vehicle is stationary. It's important to note that in this study, the Real-Time Kinematic (RTK) functionality of the GPS is never used and only differential functionality is utilized as a comparison experiment with the LIDAR SLAM based path following experiment. LIDAR is used for both localization with SLAM and perception. It is a 16 channel Velodyne LIDAR PUCK (VLP-16) as shown in Figure 18, which is mounted on the top of the vehicle horizontally to guarantee a horizontal Field of View (FOV) of 360 degrees with vertical FOV of 30 degree from the surrounding environment. A 3D point cloud is generated at a frequency of 10 Hz. Theoretically, the LIDAR's maximum detection range can reach up to

100 m depending on application while in this work, detection range used for localization was set to 80 m to achieve satisfactory point cloud density and quality.

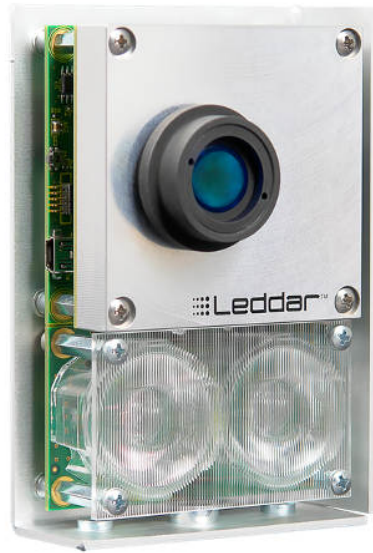

Figure 17 Leddar sensor

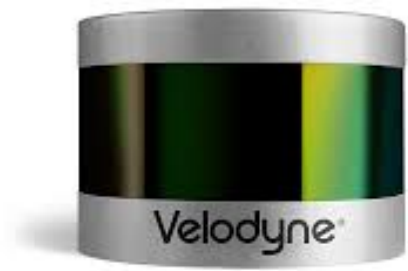

Figure 18 Velodyne LIDAR PUCK (VLP-16)

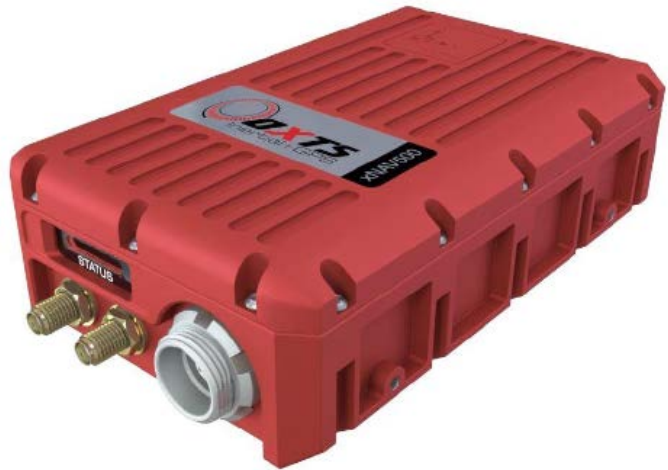

Figure 19 OXTS XNAV550 GPS

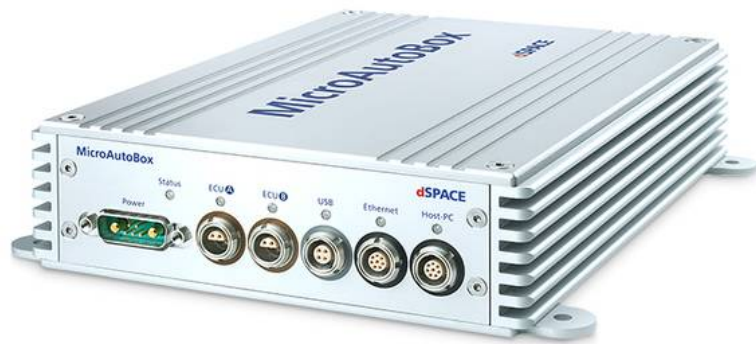

Figure 20 dSPACE Microautobox (MABx) electronic control unit

The element between the actuators and sensors is the dSPACE Microautobox (MABx) electronic control unit shown in Figure 20, that is used for rapid prototyping of the low-level lateral and longitudinal direction controllers and basic decision-making algorithms created as a Simulink models. Simulink coder is used to convert the model into embedded code and the code is uploaded to the MABx device. The generated code can later be easily embedded in a series production level electronic control unit at the end of the research and development phase.

Sensors send data to the Microautobox electronic control unit with a means of communication specific to the sensor, like CAN or User Datagram Protocol (UDP) for most of our sensors. This data is fed to controllers running within the device. Controllers are created in the Simulink and outputs of the controllers are connected to output blocks that correspond to I/O ports of the Microautobox. These I/O ports are physically connected to actuators or drivers of actuators to provide reference signal and achieve autonomous driving. The experimental vehicle also has a Dedicated Short Range Communication (DSRC) modem to communicate with other vehicles, infrastructure and pedestrians with DSRC enabled smartphones. For V2X communication, all messages are sent using the standard messages of the Society of Automotive Engineers (SAE) J2735 DSRC Message Set and use the standard communication rate of 10Hz. Devices and actuators are powered through a 12V battery placed in the trunk of the vehicle. The overall configuration of the platform discussed in this section is shown in Figure 21. The overall flow diagram of our platform is depicted in Figure 22.

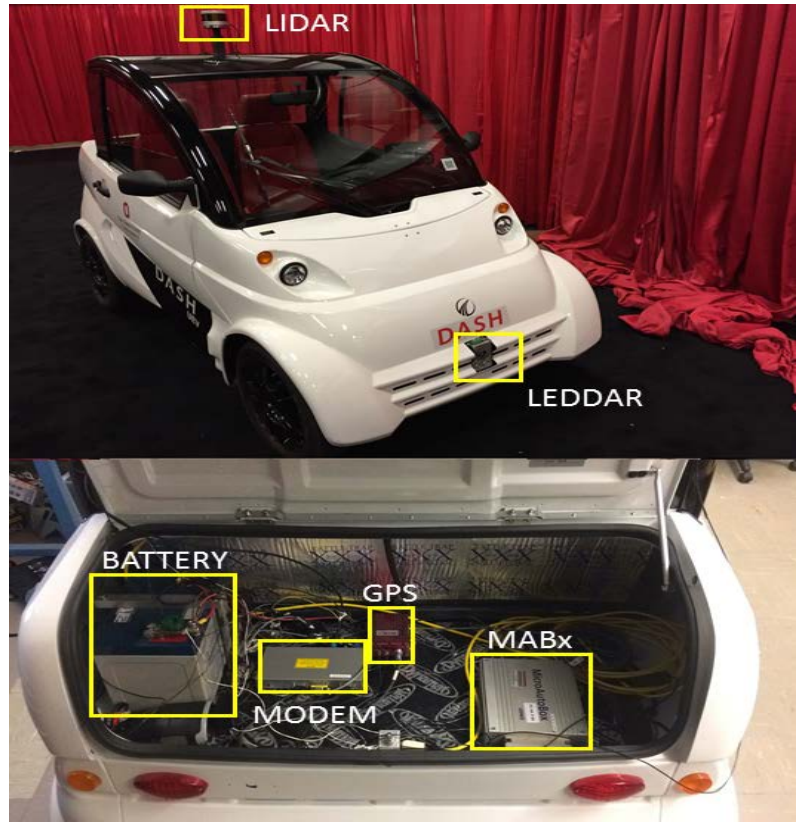

Figure 21: Hardware on the vehicle

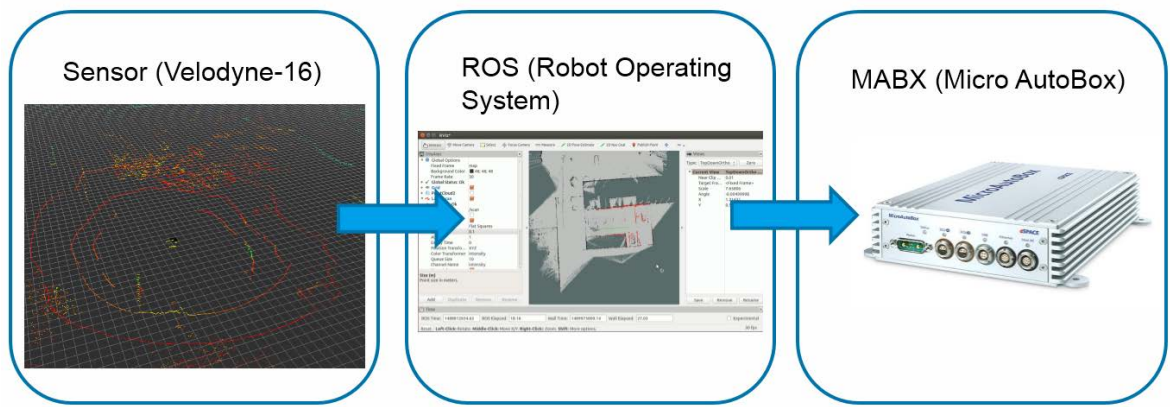

Figure 22 Flow diagram of our framework

3.4 Simulation Model and Parameters

The overall block diagram of simulation model was built up in Simulink as shown in Figure 23. The whole configuration can be generally divided into 4 parts. First is the input from sensor data (shown in Figure 24), which in this work is the localization estimation from the SLAM algorithm. Since the SLAM framework was implemented and deployed on ROS, the communication between ROS with our micro-controller MABX was realized through UDP communication developed on our own in C++. Second is the path following algorithm block (shown in Figure 25), which was involved with lateral steering control and longitudinal acceleration speed control. The estimated localization information was fed into this controller for feedback robust PID control so as to correct the deviation of the unmanned vehicle from our desired path. Third is the controller command filtering block (shown in Figure 26), which only allows 25% of the command from the above path following control algorithm block to be executed in the next stage. Fourth is the actuator block (shown in Figure 27), which is utilized for the MABX to interface with actuators. The actuator block depends on the actuator interface. Some of them are CAN blocks (brake in Dash vehicle) and some of them can be analog signal blocks (responsible for steering and throttle in Dash vehicle). The mechanical components such as throttle and brake are then executed according to the command from the actuator block to take actions such as deceleration, acceleration and steering. In this model, the empirical parameters of the vehicle are listed in Table 1.

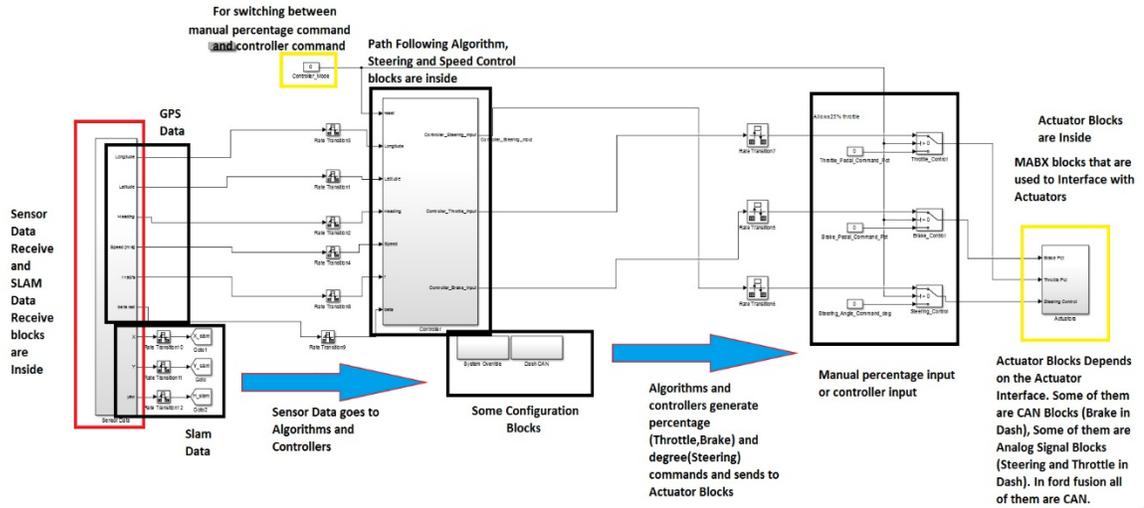

Figure 23 Overall simulation block diagram

| Parameter | Value |
| --- | --- |
| Mass (with 4 passengers) | $m = 2000 \ [kg]$ |
| Inertia moment around z axis | $J_z = 3728 \ [kg \cdot m^2]$ |
| Mass center to front of the car | $a = 1.3008 \ [m]$ |
| Mass center to rear of the car | $b = 1.54527 \ [m]$ |
| Radius of front wheel | $R_{front} = 0.3225 \ [m]$ |
| Radius of rear wheel | $R_{rear} = 0.3225 \ [m]$ |
| Height of mass center to ground | $H_g = 0.551 \ [m]$ |
| Acceleration of gravity | $g = 9.81 \ [m/s^2]$ |
| Cornering stiffness of front tire | $C_f = 1.9e5 \ [N/rad]$ |
| Cornering stiffness of rear tire | $C_r = 5e5 \ [N/rad]$ |
| Preview distance | $l_s = 2 \ [m]$ |

Table 1 Parameters of vehicle model

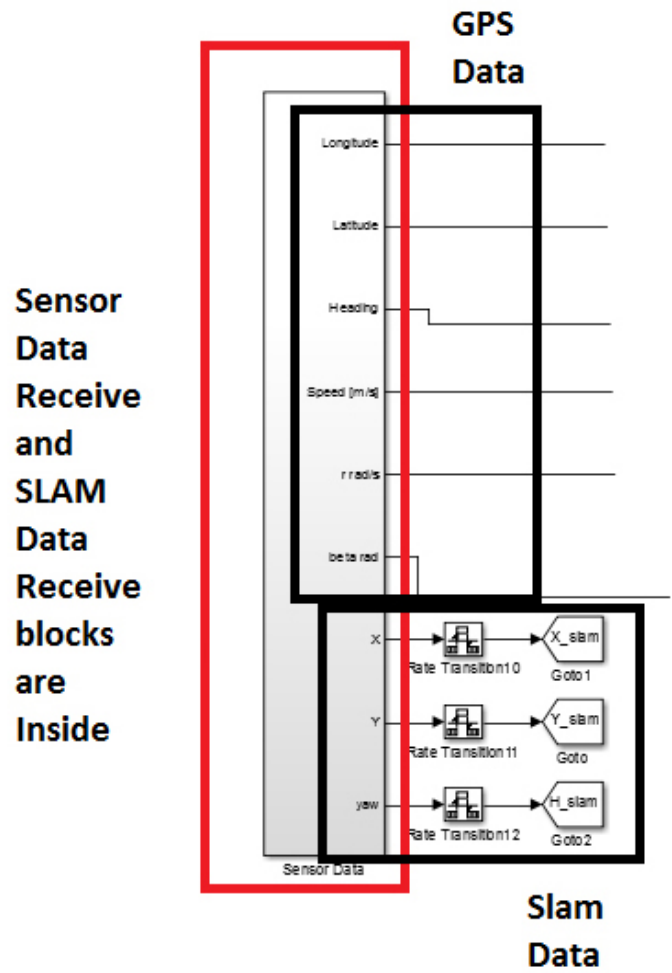

Figure 24 Sensor data block diagram

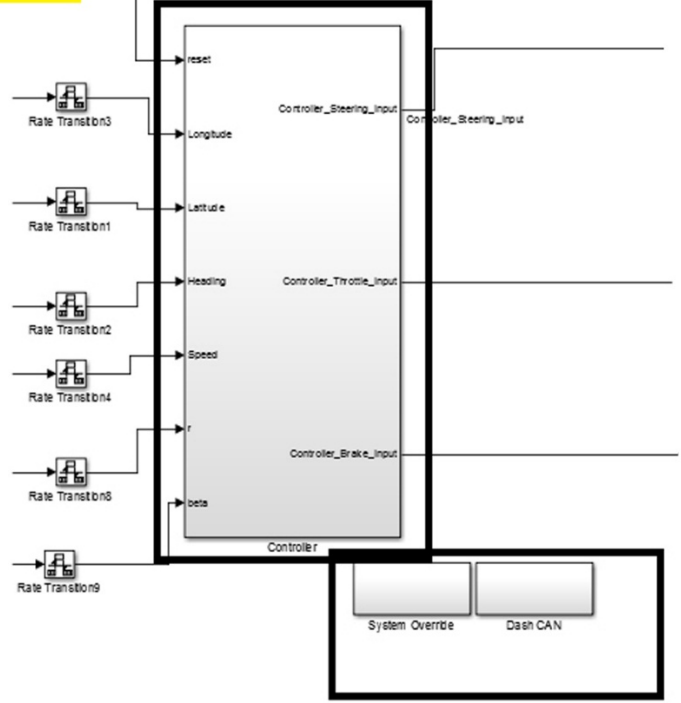

Figure 25 Block diagram of path following control algorithm

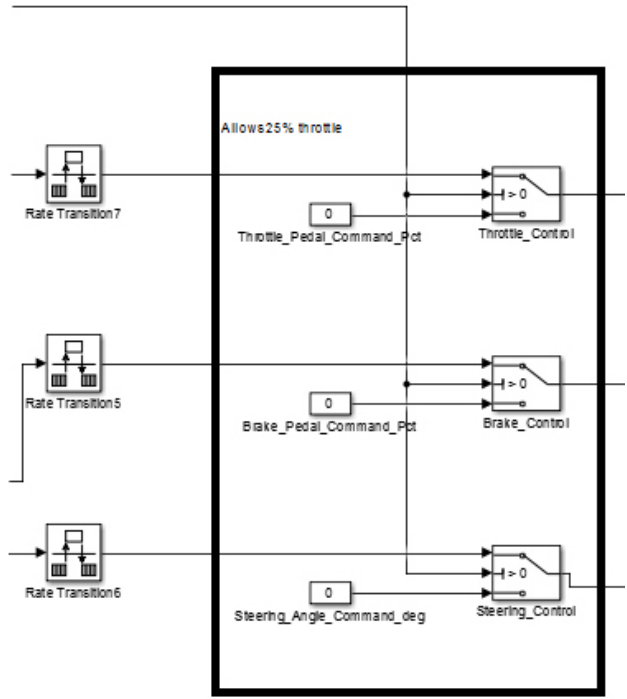

Figure 26 Block diagram of controller command filtering

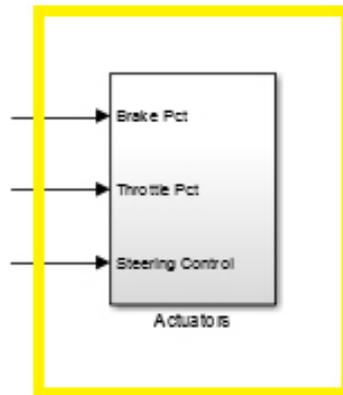

Figure 27 Block diagram of actuator interface

Chapter 4. Results and Evaluations

We conducted extensive experimental validations of our system including offline SLAM system test on collected data as well as real time field experiment in the area around the initial autonomous vehicle (AV) pilot test route, a small segment in an underserved area of campus designated by The Ohio State University, as shown in Figure 28. All the algorithms relevant to LIDAR data processing and SLAM as described above are implemented in C++ because of its efficiency of real time performance. Performances are evaluated between the SLAM system proposed in [9] and the extended version proposed in this paper. Traditional path following experiment result based on high accuracy GPS similar to the previous work is compared with this innovative SLAM based path following experiment result, demonstrating the feasibility and effectiveness of this compounded system. Note that randomness is inevitably introduced by probabilistic occupancy grid map model in the SLAM system. For this reason, the experiment results are reported based on the median performance of several runs.

Real time SLAM algorithm is carried out with an I7-6700HQ (8 cores @ 2.60 GHz) and 4Gb RAM on the Robot Operating System (ROS) [19], an open source operating system providing services designed for heterogeneous computer cluster in Linux environment. User Datagram Protocol (UDP) communication is built up between ROS and MABx for localization information transfer. Regional localization information delivered by SLAM algorithm is sent to MABx for further decision making and control strategy, e.g. longitudinal or lateral control.

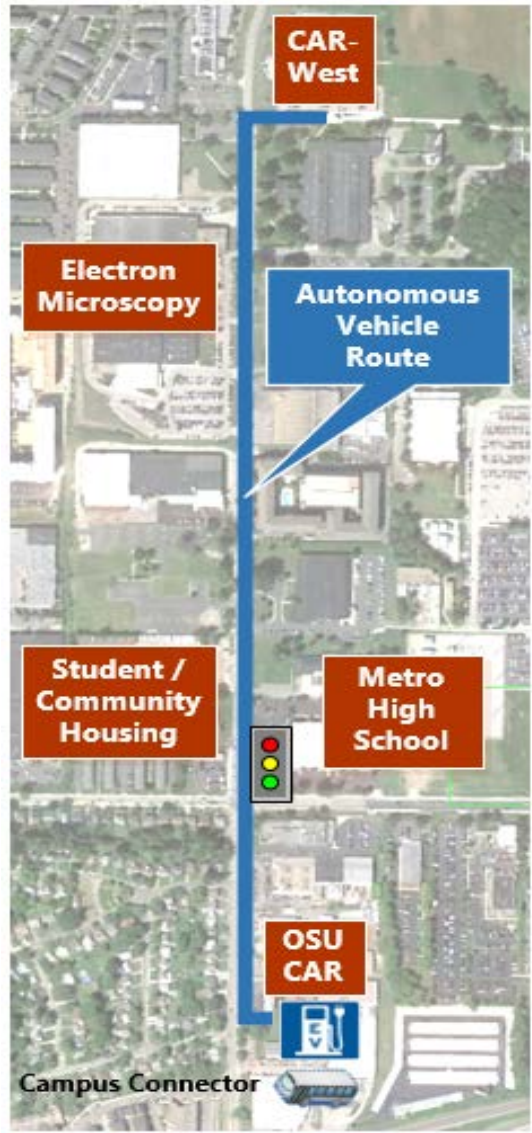

Figure 28 Autonomous vehicle test route from Car-West to Car (scale 1:8000)

4.1 SLAM Offline Simulation Evaluation

In order to quantitatively evaluate our proposed SLAM system against Hector SLAM, both SLAM systems are tested on the same LIDAR data collected around our lab, Car-West. Due to the absence of "ground truth", alignment error yielded in both algorithms is reported for comparison. Ideally, with sufficient accuracy, the alignment error (described in equation (11)) should be very small. However, inevitably introduced sensor noise and non-smooth approximation of the optimization model make the solution of pose estimation only able to approach real pose but never perfectly equivalent and hence total alignment error always exists. Therefore, in the same context, the smaller the alignment error, the higher the accuracy that is achieved and hereby we evaluate the performance by comparing their alignment error and iterations implemented in each alignment, which can reflect their estimation accuracy as well as their convergence speed. Considering that offline SLAM accuracy is similar to its real time accuracy, this comparison can effectively validate the overall performance of our proposed SLAM system against the Hector SLAM.

The ultimate map generated by our proposed SLAM system is overlapped with the same location obtained from Google Earth for comparison convenience as shown in Figure 29, where the map generated by our proposed SLAM is in shadow and red line is the test trajectory. It is important to note that the map from Google Earth is not strictly top down view. Thus here a minor shift is necessarily used to keep the edges of the mapped buildings consistent with their actual corresponding edges in Google Earth. In this experiment, raw LIDAR data is initially collected by VLP-16 along the test trajectory which starts from the backyard of Car-West, passing through an open field which is sufficiently challenging because of the limited landscapes for matching alignment and textureless wall. Another challenging part of this test trajectory is a sharp 180 degree turn in the front of the parking lot of the lab building, which demands fast convergence and robustness of the nonlinear optimization model implemented in the SLAM system.

Figure 30 shows both complete and regional localization estimation from the two SLAM systems along the test trajectory. The smoother localization given by our proposed SLAM system with the integrated automated drive control systems can dramatically improve passenger comfort while taking a ride in the shuttle. Table I illustrates the average alignment error and average iteration steps required between the two SLAM systems. It can be clearly observed that in some runs, our proposed SLAM can effectively reduce the alignment error to a relatively lower level despite the fact that in almost half of the runs the benefit is not distinct. Results of the average alignment error from Figure 34 can further prove this property. This can be attributed to the defect of this optimization based SLAM system where global minimum cannot be guaranteed and scan end point outliers can inevitably introduce noise to the system. Therefore, a reliable preprocessing model of the scan end points is desired as an extension to this framework, which may be an interesting topic in future work. Although in our proposed SLAM system additional iteration steps are sacrificed for better alignment compared with Hector SLAM, in which the iteration step is set to a fixed value and naturally convergence cannot be guaranteed, the increased iteration step is still in an acceptable range for real time performance according to our real-time experiments.

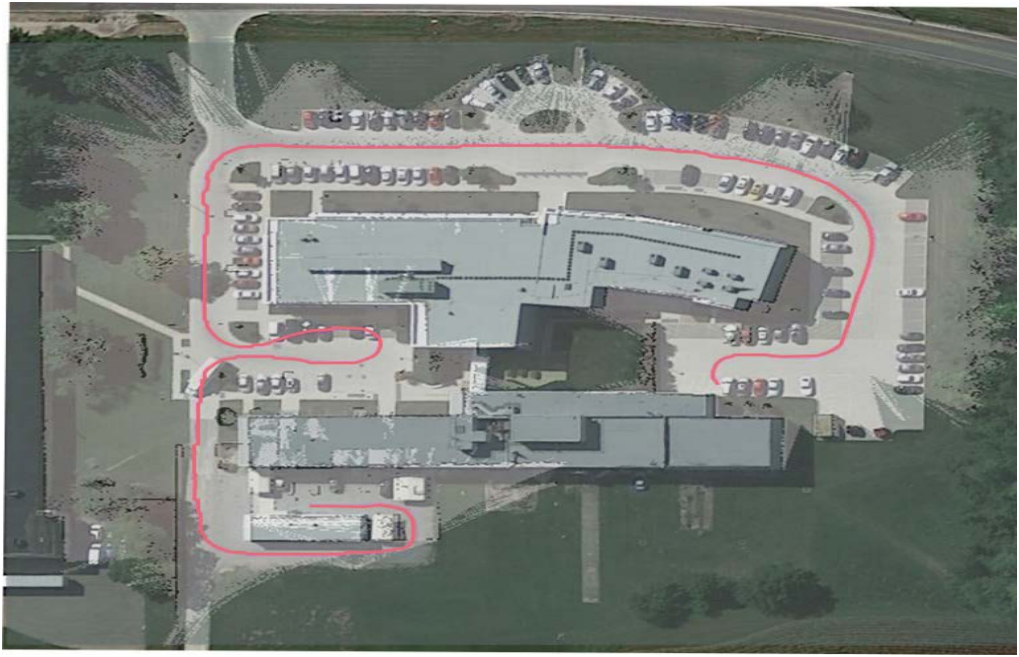

Figure 29 Generated map overlapped with Google Earth

|  | Proposed SLAM | Hector SLAM |
|---|---|---|
| Average alignment error | 78.759 | 84.107 |
| Average iteration step | 6.557 | 3.400 |

Table 2 Performance comparison between our proposed SLAM with Hector SLAM. Alignment error is accumulated error of occupancy value, which is dimensionless

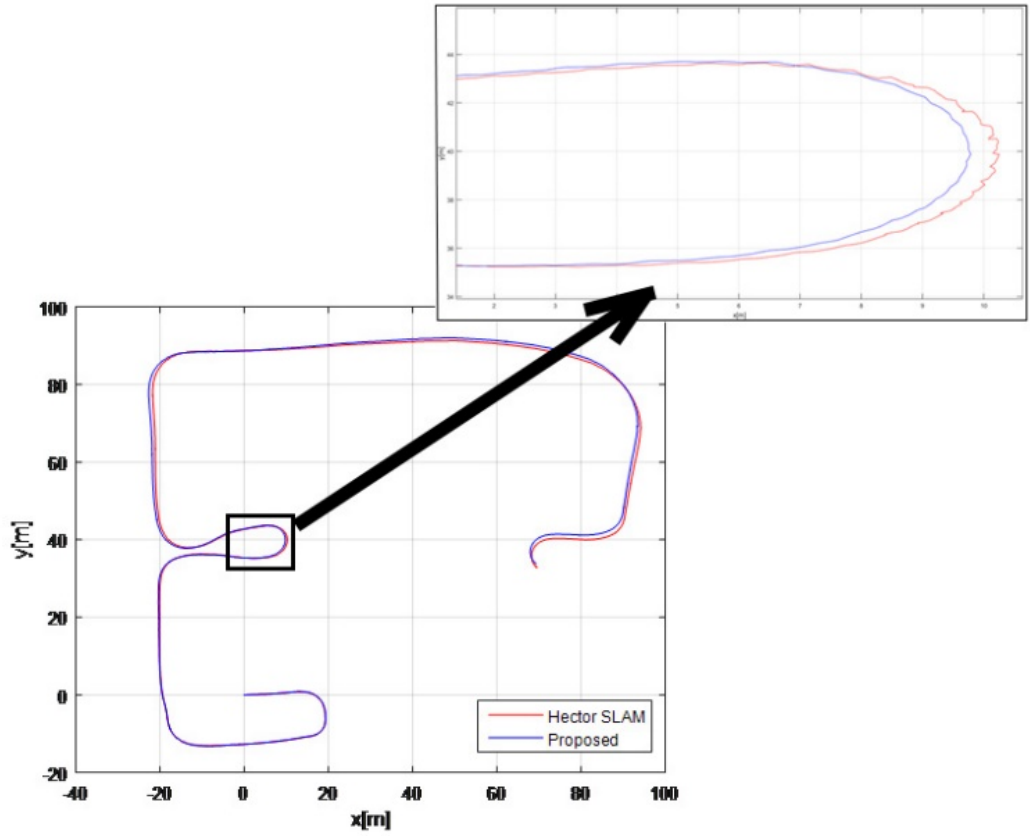

Figure 30 Trajectory comparison between our proposed SLAM (blue) with Hector SLAM (red)

## 4.2 Real Time Path Following Performance

In addition to quantitative evaluation of our proposed SLAM system, various real world experiments are also conducted to validate its feasibility and adaptivity of integration with the control system. We first manually drive the shuttle along the pre-determined trajectory around our lab building, as shown in Figure 31, to collect GPS points, from which the desired path is then generated for path following reference.

Figure 32 and Figure 33 show the actual path following trajectory performed by our proposed SLAM system and RTK GPS separately compared with the desired path. The coordinate of starting position is set to the origin in the following plots for comparison convenience. It can be observed that similar to GPS, SLAM based path following can be achieved comparable to GPS based result, though with occasional minor error, which again proved the supplemental functionality of our proposed SLAM system in GPS not accessible cases. Figure 34 shows the root-mean-square error (RMSE) along the whole path following trajectory performed by SLAM compared with the same experiment setting but performed by differential GPS. The shuttle speed of both path following approaches are kept at an average value of 12 km/h. As can be seen from the experimental results, conventional path following that relies on highly accurate differential GPS has the expected performance with appropriate lateral controller design. The overall performance of GPS is better than SLAM, but SLAM based path following tends to have even smaller RMSE at some regions, e.g. at points of $0.7 \times 10^5, 1.5 \times 10^5, 1.8 \times 10^5$ which are at the corners of the trajectory. The fact suggests that this SLAM system can provide precise estimation of the shuttle orientation while there may exist some delay or inaccuracy in the orientation angle provided by differential GPS, which is computed based on compass. It demonstrates that localization and perception system that purely relies on LIDAR can supplement the cases when GPS is not available or a lower cost and hence lower accuracy GPS is desirable for intelligent shuttles.

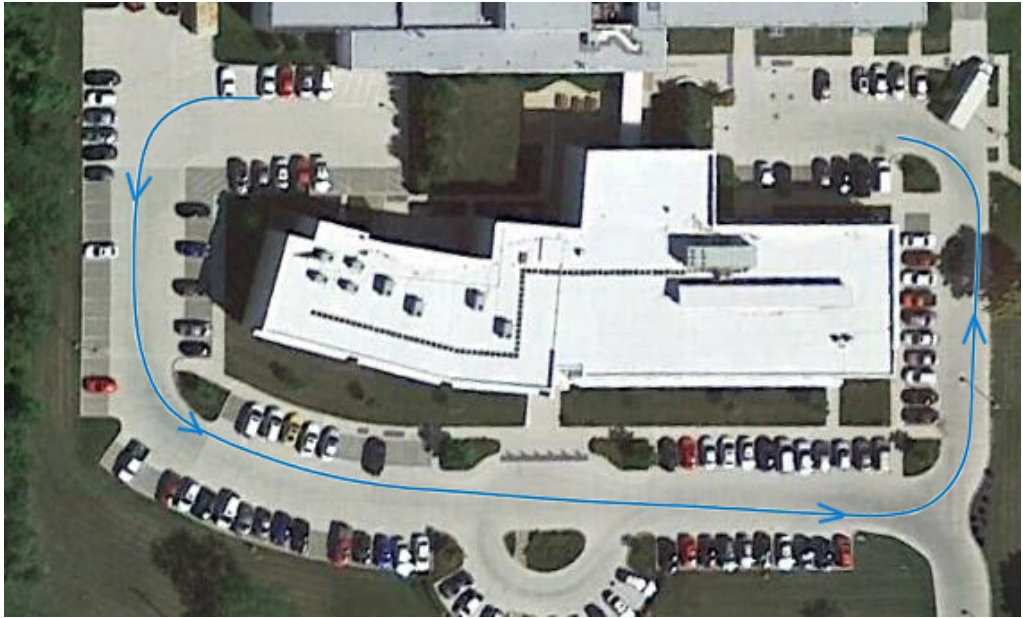

Figure 31 Trajectory on satellite image

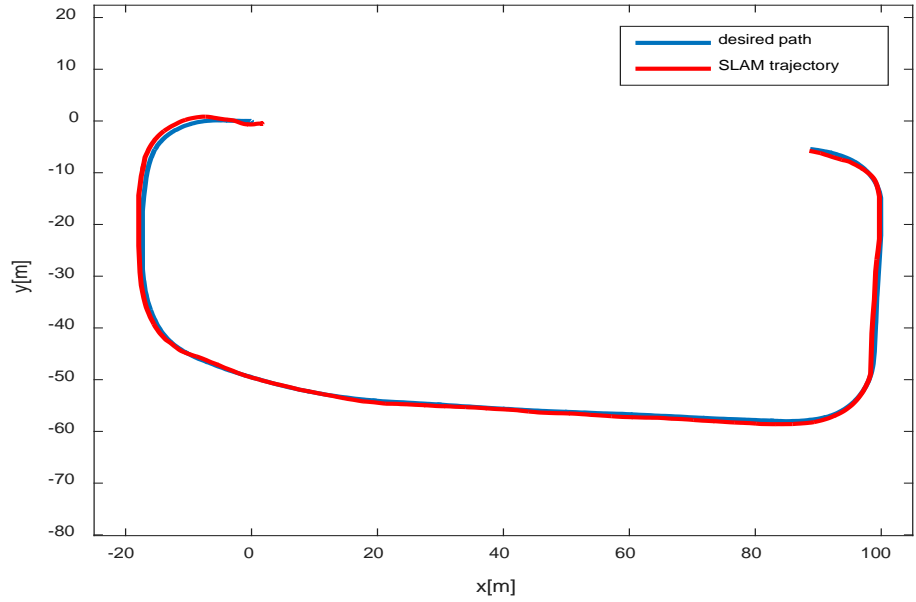

Figure 32 Desired path compared to our proposed SLAM path following trajectory

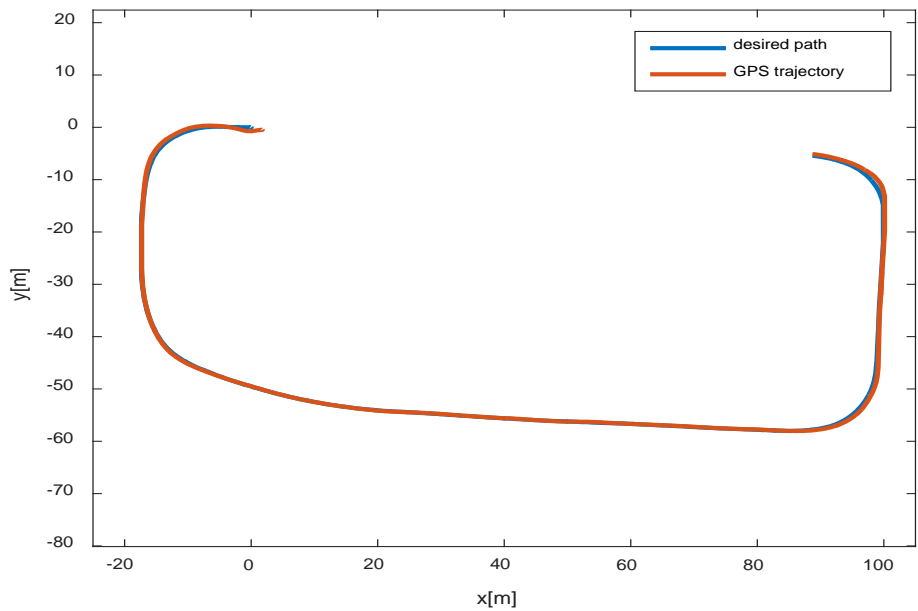

Figure 33 Desired path compared to GPS path following trajectory

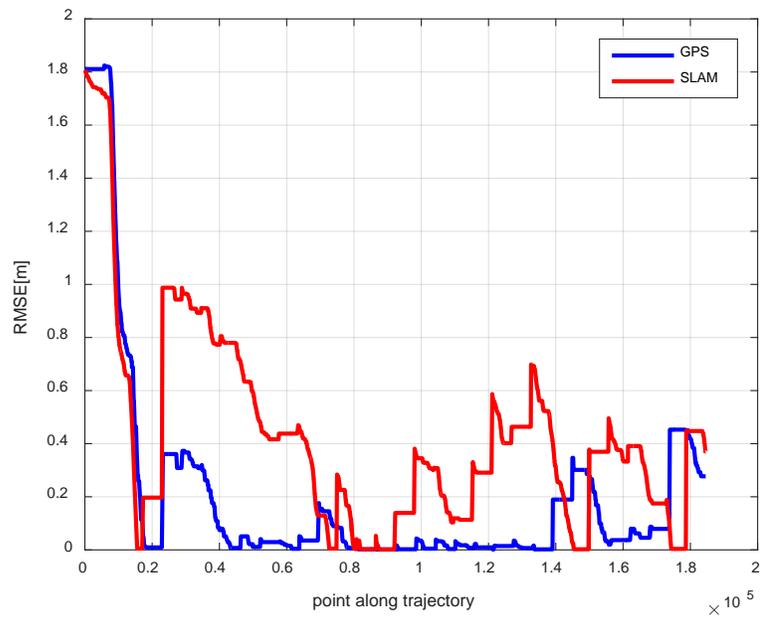

Figure 34 RMSE in lateral direction comparison between our proposed SLAM based path following and GPS based path following

Chapter 5. Conclusions

This thesis presented preliminary work for a AV shuttle deployment in the AV pilot test route of the Ohio State University. GPS and LIDAR SLAM are both used for localization and path generation. Since GPS based localization and path following was presented in our earlier work, this paper concentrated on a LIDAR SLAM system which is inherited from the Hector SLAM framework and based on the Levenberg-Marquardt algorithm. It was demonstrated that this LIDAR SLAM algorithm can be used for self-localization of our low speed autonomous shuttle. Extensive experiments were conducted for offline SLAM performance evaluation as well as real world experiments for path following in a parking lot for safety. The proposed SLAM system was compared with the state of art 2D SLAM approach especially in terms of scan alignment accuracy and seen to provide dynamically reasonable pose estimation. As a pre-requisite to testing autonomous driving on the actual AV pilot test route, this route was replicated in our HiL simulator for developing and testing low level controllers and decision making logic. GPS and Leddar sensors, traffic and the traffic light are ale to be emulated in the HiL simulator while the low level control ECU and the DSRC radios used for V2I and V2V communication were real hardware. LIDAR sensor emulation work is in progress and will allow us to implement LIDAR based algorithms for both localization, e.g. SLAM, and obstacle detection and classification within the HiL simulator.